\UseRawInputEncoding
\documentclass[preprint,12pt]{elsarticle}
\usepackage{amssymb}
\usepackage{placeins}
\usepackage{rotating}
\usepackage{colortbl}
\usepackage{lscape}
\usepackage{longtable}
\usepackage{adjustbox}
\usepackage{pdflscape}
\usepackage{algorithm}
\usepackage{algorithmicx}
\usepackage{algpseudocode}
\usepackage{multirow}
\usepackage{booktabs}
\usepackage{xcolor}
\usepackage{makecell}
\definecolor{lightgray}{gray}{0.95}
\definecolor{highlightrow}{rgb}{0.88,0.88,0.88}
\newcolumntype{L}{>{\centering\arraybackslash}m{6cm}}
\newcolumntype{M}{>{\centering\arraybackslash}m{2.5cm}l}
\biboptions{sort&compress}

\journal{Neurocomputing}

\begin{document}

\begin{frontmatter}

\title{Decoding Neural Emotion Patterns through Large Language Model Embeddings}
\author[inst1]{Gideon Vos}

\affiliation[inst1]{organization={College of Science and Engineering, James Cook University},
            addressline={James Cook Dr}, 
            city={Townsville},
            postcode={4811}, 
            state={QLD},
            country={Australia}}

\author[inst1]{Maryam Ebrahimpour}
\author[inst2]{Liza van Eijk}
\author[inst3]{Zoltan Sarnyai}
\author[inst1]{Mostafa Rahimi Azghadi}

\affiliation[inst2]{organization={College of Health Care Sciences, James Cook University},
            addressline={James Cook Dr}, 
            city={Townsville},
            postcode={4811}, 
            state={QLD},
            country={Australia}}

\affiliation[inst3]{organization={College of Public Health, Medical, and Vet Sciences, James Cook University},
            addressline={James Cook Dr}, 
            city={Townsville},
            postcode={4811}, 
            state={QLD},
            country={Australia}}
\begin{abstract}

\noindent Understanding how emotional expression in language relates to brain function remains a key challenge in neuroscience. Traditional neuroimaging provides valuable insight but is costly and limited to controlled laboratory settings. Here, we present a computational framework that explores potential links between emotional content in natural language to neuro-anatomical regions associated with affective processing. This, when validated through complementary neuroimaging, may enable scalable, imaging-free investigation of emotion-brain relationships. Our approach combines text embeddings, dimensionality reduction, and clustering to identify emotional states, which are then mapped to relevant brain regions. The framework was evaluated across three applications: (i) comparing healthy and depressed individuals, (ii) analyzing a large-scale emotion dataset, and (iii) contrasting human and large language model (LLM) outputs. Emotion intensity was quantified using a lexical scoring system sensitive to keywords, syntax, and modifiers, producing computationally plausible emotion-to-region clusters with visualization mapping. Across experiments, the framework distinguished healthy from depressed participants through distinct computational activation patterns and revealed systematic differences between human and LLM-generated texts in predicted computational engagement. A key finding emerged: depressed individuals exhibited reduced emotional diversity, showing 2.2 - 2.7 times more homogeneous emotional expression than healthy controls, suggesting that emotional rigidity may serve as a computational marker of depression. These computational patterns represent testable hypotheses, requiring further validation through neuroimaging. This work establishes a scalable, cost-effective tool for advancing both clinical and computational models of emotion, and provides a neuro-inspired benchmark for assessing how closely AI-generated language mirrors human emotional expression.

\end{abstract}

\begin{graphicalabstract}
\begin{center}
  \makebox[\textwidth]{\includegraphics[width=\paperwidth]{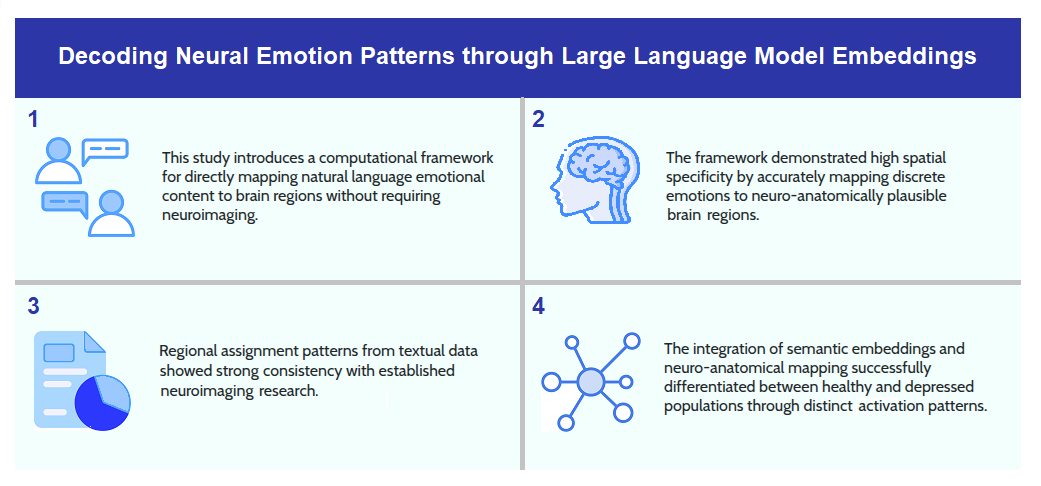}}
\end{center}
\end{graphicalabstract}

\begin{highlights}
\item This study introduces a computational framework for directly mapping natural language emotional content to brain regions without requiring neuroimaging.
\item The integration of embeddings and neuro-anatomical mapping successfully differentiated between healthy and depressed populations through distinct activation patterns.
\item The framework demonstrated high spatial specificity by accurately mapping discrete emotions to neuro-anatomically plausible brain regions.
\item  Regional assignment patterns were derived from established neuroimaging coordinates, creating computationally-predicted activation patterns that require independent validation.
\item In favor of reproducible research and to advance the field, all programming code used in this study is made publicly available. 
\end{highlights}


\end{frontmatter}



\section{Introduction}

\noindent Understanding the neural correlates of emotion has traditionally relied on neuroimaging modalities such as electroencephalography (EEG) and functional magnetic resonance imaging (fMRI) \cite{lindquist2012, vytal2010, murphy2003}. These approaches have identified key regions such as the amygdala, insula, anterior cingulate cortex, and prefrontal cortex as central to emotional processing \cite{saarimaki2022, phillips2003, sliz2012, ibrahim2022}. However, neuroimaging studies face substantial challenges including high cost, limited accessibility, controlled laboratory constraints, and reduced ecological validity when studying naturalistic emotion \cite{drevets2001, hastings2004}. Consequently, there is growing interest in computational alternatives capable of mapping emotion-brain relationships using naturalistic data sources such as text. \\

\noindent Recent advances in language modeling have shown striking parallels between large language model (LLM) embeddings and human brain activity. Caucheteux \emph{et al.} \cite{caucheteux2022} demonstrated that pre-trained language models align with neural responses without task-specific training, while Toneva \emph{et al.} \cite{toneva2022} and Schrimpf \emph{et al.} \cite{schrimpf2021} confirmed geometric and predictive correspondence between LLM-derived representations and cortical activation patterns during language comprehension. These findings suggest that the distributed semantic representations learned by LLMs may approximate aspects of neural language encoding in a computational sense, providing a promising foundation for computationally modeling potential brain–language relationships. Similar to recent advances in visual recognition that emphasize non-parametric, embedding-based reasoning, our approach relies on representational geometry rather than parametric classification. The Deep Nearest Centroids (DNC) framework \cite{Wang2022DeepCentroids}, for instance, achieves interpretable decision-making by associating test samples with class sub-centroids in embedding space, enabling both explainability and cross-domain transferability. This methodological alignment underscores the potential of embedding-based representations for interpretable, model-derived approximations of language-brain relationships. \\

\noindent Parallel work has explored how emotionally-charged language engages distinct neural systems. Tomasino \emph{et al.} \cite{tomasino2023} and Chen \emph{et al.} \cite{chen2023} showed that linguistic valence correlates with differential activation of prefrontal and limbic regions, aligning with meta-analytic findings that positive and negative emotions recruit left and right hemispheric structures, respectively \cite{davidson2004}. Zhou \emph{et al.} \cite{zhou2022} and Xiao \emph{et al.} \cite{xiao2021} further demonstrated that embeddings derived from emotional text can be linked to fMRI and EEG signals, highlighting distributed yet consistent mappings across emotion categories. Together, these studies underscore that emotional semantics in language may provide a viable proxy for corresponding neural patterns. \\

\noindent Beyond theoretical mapping, embedding-based models may capture clinically relevant differences in emotional and linguistic processing. Individuals with depression and related conditions exhibit altered word choice, affective tone, and syntactic complexity \cite{campbell2004, brosch2022, liu2025, ge2025, lorenzoni2025, zhong2025}. If such language deviations correspond to altered brain activation patterns, computational emotion-brain mapping could enable scalable biomarkers for mental health \cite{ramirez2022, schwartz2014, zhou2021}. Relatedly, these methods may help distinguish between human and machine-generated text through their inferred emotional brain activation patterns \cite{bulat2023}. \\

\noindent Despite prior progress, there is no existing imaging-free framework that directly links natural language emotion to specific neuro-anatomical regions. The present study introduces a computational approach that transforms emotional embeddings into brain-region activation patterns. Specifically, we aim to:
\begin{itemize}
    \item Develop a novel computational framework for mapping emotional language to brain region coordinates derived from neuroimaging literature.
    \item Generate testable hypotheses by applying this framework to differentiate between healthy and depressed populations.
    \item Produce emotion-region association patterns as predictions requiring orthogonal validation through independent neuroimaging studies.
\end{itemize}
This computational approach offers a scalable, cost-effective alternative to neuroimaging, enabling interpretable emotion–brain mapping from text alone.

\section{Methods}

\subsection{Datasets}

\noindent Three text-based datasets were employed in this study (Table \ref{tab:datasets}). The DIAC-WOZ dataset \cite{dataset_diac} comprises annotated interview transcripts from individuals diagnosed with depression and healthy controls. The GoEmotions dataset \cite{dataset_emotions} includes 58,000 Reddit \cite{reddit} comments manually labeled into 27 emotion categories (or neutral). The Schema-Guided Dialogue dataset \cite{dataset_google} represents nearly half a million sentences comprised of human and LLM chatbot interactions. All datasets consist of texts produced by native English speakers. 

\begin{table}[!h]
\centering
\caption{\label{tab:datasets}Datasets utilized in this study.}
\resizebox{\textwidth}{!}{
\begin{tabular}{lll}
\hline\hline
\textbf{Dataset}  & \textbf{Emotions} & \textbf{Subjects}                                         \\
\hline
DAICWOZ\cite{dataset_diac}  & Healthy and Depressed Categories                          & 134 Clinical interview transcripts  \\    
GoEmotions \cite{dataset_emotions}                              & 27 Emotion Categories                           & 58k English Reddit comments  \\
The Schema-Guided Dialogue Dataset \cite{dataset_google} & Human and Chat bot conversations & 463,282 English sentences \\
\hline
\end{tabular}
}
\end{table}
\FloatBarrier

\subsection{Text Preprocessing and Embedding Generation}

\begin{figure}[h!]
\centering
\includegraphics[height=0.9\textheight]{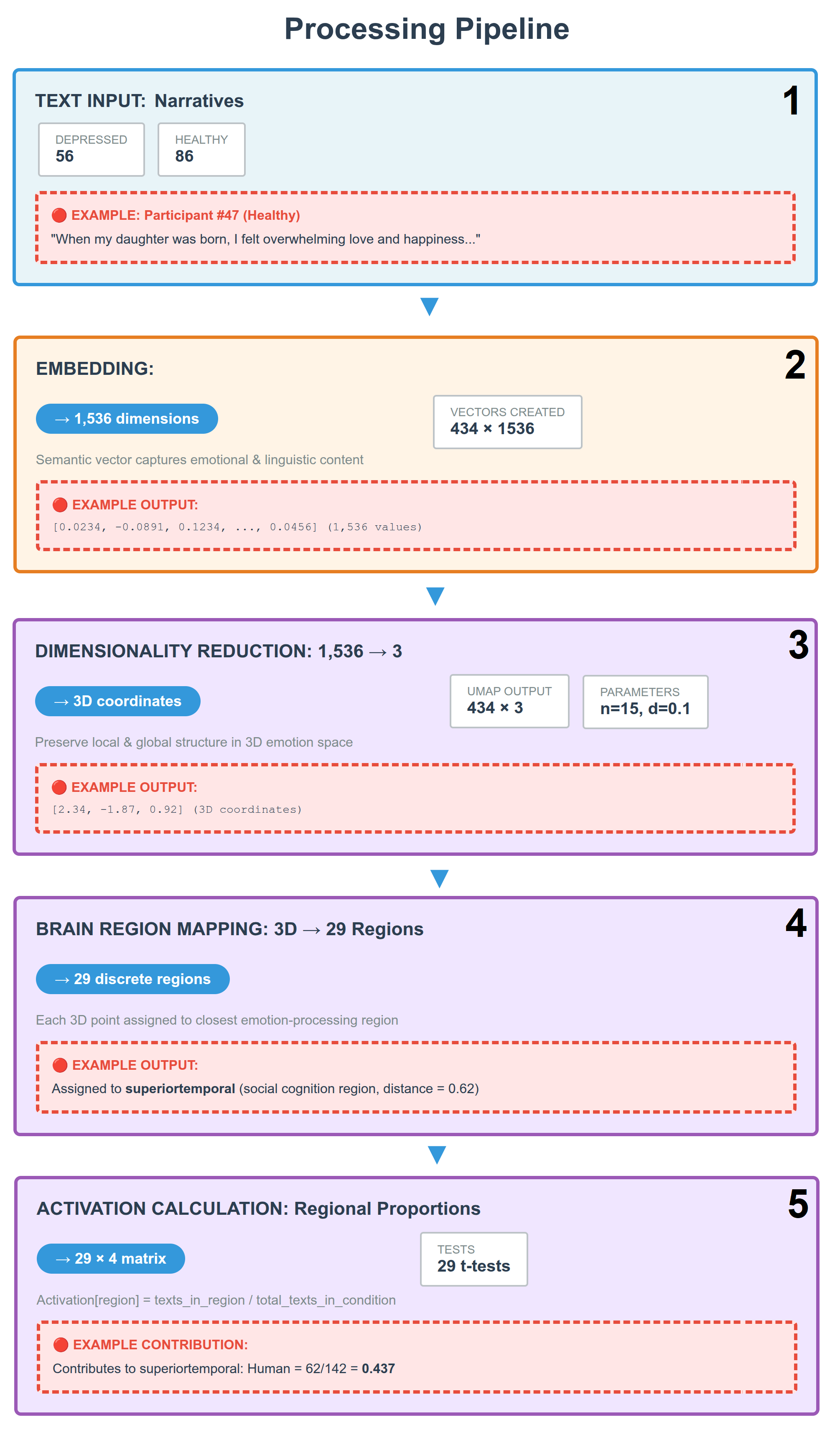}
\caption{\label{fig:figure1} Five-step computational pipeline to convert natural language text to embeddings, reduce dimensionality, cluster to emotional groups, map to brain regions and calculate activations.}%
\end{figure}

\noindent Texts were divided into 300 character segments using sentence boundaries. Each segment was converted into a 1,536-dimensional vector using OpenAI's \emph{text-embedding-ada-002} model \cite{openai} (Figure \ref{fig:figure1}, Step 2). This model was selected for its well-validated emotional and semantic coverage \cite{liu2025, ge2025, lorenzoni2025, zhong2025}, avoiding bias that might arise from custom embeddings. The programming pseudo-code for steps 1 and 2 of Fig. \ref{fig:figure1} is shown in Algorithm \ref{alg:preprocessing_embeddings}.

\FloatBarrier

\begin{algorithm}
\caption{Text Preprocessing and Embedding Generation}
\label{alg:preprocessing_embeddings}

\textbf{Input:} Text datasets $D = \{d_1, d_2, d_3\}$ \\
\textbf{Output:} 1536-dimensional embeddings matrix

\begin{algorithmic}[1]

\State \textbf{// Step 1: Text Preprocessing and Chunking}
\Function{PreprocessTexts}{$texts$}
    \State $chunks \leftarrow []$
    \For{each $text$ in $texts$}
        \State $segments \leftarrow$ split $text$ into $\approx$300 character chunks using periods
        \State $chunks$.append($segments$)
    \EndFor
    \State \textbf{return} $chunks$
\EndFunction

\State
\State \textbf{// Step 2: Text Embedding Generation}
\Function{GetAdaEmbeddings}{$texts$}
    \State Initialize OpenAI client with API key
    \State $embeddings \leftarrow []$, $batch\_size \leftarrow 2000$
    \For{$i = 0$ to $len(texts)$ step $batch\_size$}
        \State $batch \leftarrow texts[i:i+batch\_size]$
        \State $response \leftarrow$ client.embeddings.create(model="text-embedding-ada-002", input=$batch$)
        \State $batch\_embeddings \leftarrow$ extract embeddings from $response$
        \State $embeddings$.extend($batch\_embeddings$)
    \EndFor
    \State \textbf{return} $np.array(embeddings)$ \textit{// Shape: (n\_samples, 1536)}
\EndFunction

\end{algorithmic}
\end{algorithm}

\subsection{Dimensionality Reduction and Spatial Mapping}

\noindent The high-dimensional embeddings underwent a dimensionality reduction process (Figure \ref{fig:figure1}, step 3) using Principal Component Analysis (PCA) to reduce the dimensionality to three components, representing the minimum number of dimensions required for spatial brain mapping. As PCA was used primarily to obtain a spatially interpretable representation for brain-region visualization, reducing embeddings to three components captures only a small portion of the total variance. Clustering performed within this 3D space therefore reflects an intentional interpretability trade-off rather than an assumption that these components preserve most of the embedding structure. The choice of a 3D subspace was made to enable direct mapping onto MNI coordinates and to support the cortical surface visualizations, while acknowledging that clustering in a low-variance space may limit the capture of finer-grained embedding structure.

\subsection{Emotional Intensity Estimation}

\noindent Emotional intensity was computed through a lexicon-based scoring scheme combined with syntactic modifiers (Figure \ref{fig:figure1}, Step 3). Words were assigned base intensities (mild 0.3, moderate 0.6, high 0.8, extreme 1.0), adjusted by amplifiers (very, really 0.3), absolutists (always, never 0.2), and punctuation cues (0.25, 0.15). Uppercase text added 0.5; all scores were capped at 2.0. This weighting follows continuous affect-intensity principles from established lexica (NRC \cite{mohammad2018word}, ANEW \cite{bradley1999anew}) rather than empirically tuned parameters. Algorithm 2 outlines the combined dimensionality reduction, spatial mapping and intensity estimation steps denoted in Figure \ref{fig:figure1} as Step 3.

\begin{algorithm}
\caption{Dimensionality Reduction and Emotional Intensity Estimation}
\label{alg:reduction_intensity}

\textbf{Input:} High-dimensional embeddings from Step 2.\\
\textbf{Output:} 3D embeddings and intensity scores

\begin{algorithmic}[1]

\State \textbf{// Step 3A: Dimensionality Reduction}
\Function{FitTransformEmbeddings}{$embeddings$}
    \State $n\_components \leftarrow \min(3, n\_samples, n\_features)$
    \State Initialize StandardScaler() and PCA($n\_components$)
    \State $embeddings\_scaled \leftarrow$ scaler.fit\_transform($embeddings$)
    \State $embeddings\_3d \leftarrow$ pca.fit\_transform($embeddings\_scaled$)
    \If{$embeddings\_3d.shape[1] < 3$}
        \State Pad with zeros to ensure 3D representation
    \EndIf
    \State \textbf{return} $embeddings\_3d$
\EndFunction

\State
\State \textbf{// Step 3B: Emotional Intensity Estimation}
\Function{EstimateEmotionIntensity}{$texts$}
    \State Define $word\_scores$: Extreme (1.0), High (0.8), Moderate (0.6), Mild (0.3)
    \State $intensities \leftarrow []$
    \For{each $text$ in $texts$}
        \State $intensity \leftarrow 0.1$, $words \leftarrow$ extract words from $text.lower()$
        \For{each $word$ in $words$}
            \State $intensity \leftarrow intensity + word\_scores.get(word, 0)$
        \EndFor
        \State Apply modifiers: +0.3 (intensifiers), +0.2 (absolutists)
        \State $intensity \leftarrow intensity + 0.25 \times \min(text.count('!'), 4)$
        \State $intensity \leftarrow intensity + 0.15 \times \min(text.count('?'), 3)$
        \If{$text.isupper()$ and $len(text) > 3$} $intensity \leftarrow intensity + 0.5$ \EndIf
        \State $intensities$.append($\min(intensity, 2.0)$)
    \EndFor
    \State \textbf{return} $np.array(intensities)$
\EndFunction
\end{algorithmic}
\end{algorithm}
\FloatBarrier

\subsection{Emotion Region Clustering}

\noindent K-means clustering was applied to the 3D PCA-transformed embeddings to identify distinct emotional patterns within the data (Figure \ref{fig:figure1}, step 4). The number of clusters was set to match 29 predefined anatomical brain regions, establishing a direct correspondence between emotional content clusters and neuro-anatomical structures \cite{phillips2003, lindquist2012, kober2008, etkin2011, seeley2007}. Of the 29 anatomically defined brain regions selected, 14 (Table \ref{tab:brainregions}) have been consistently implicated in emotion processing \cite{murphy2003, phan2002, vytal2010, lindquist2012}. 

\begin{table}[h]
\centering
\caption{\label{tab:brainregions}Predefined anatomical brain regions used for K-means clustering.}
\resizebox{\textwidth}{!}{
\begin{tabular}{|l|l|}
\hline
\textbf{Anatomical Category} & \textbf{Brain Regions} \\
\hline\hline
Frontal Lobe & Medial Orbitofrontal (bilateral), Lateral Orbitofrontal (bilateral), \\
             & Pars Opercularis (bilateral), Rostral Middle Frontal (bilateral), Superior Frontal (bilateral) \\
\hline
Temporal Lobe & Parahippocampal (bilateral), Fusiform (bilateral), Entorhinal (bilateral) \\
\hline
Cingulate Gyrus & Rostral Anterior Cingulate (bilateral), Caudal Anterior Cingulate (bilateral), Posterior Cingulate (bilateral) \\
\hline
Insula & Insula (bilateral) \\
\hline
Occipital Lobe & Lingual (bilateral), Cuneus (bilateral) \\
\hline
\end{tabular}
}
\end{table}

\FloatBarrier

\subsection{Cluster-to-Region Neuro-anatomical Mapping}

\begin{figure}[h!]
\centering
\includegraphics[width=\textwidth]{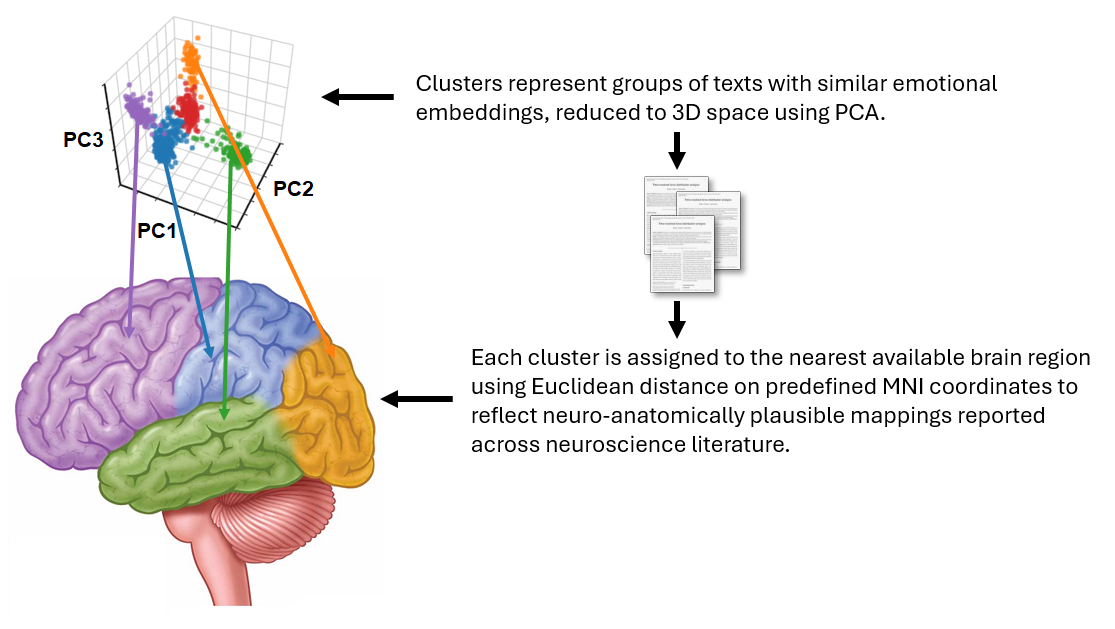}
\caption{\label{fig:figure2} Text embedding clusters mapped to brain regions via PCA dimension reduction based on neuro-scientifically plausible regions.}%
\end{figure}

\begin{figure}[h!]
\centering
\includegraphics[width=0.9\textheight, angle=90]{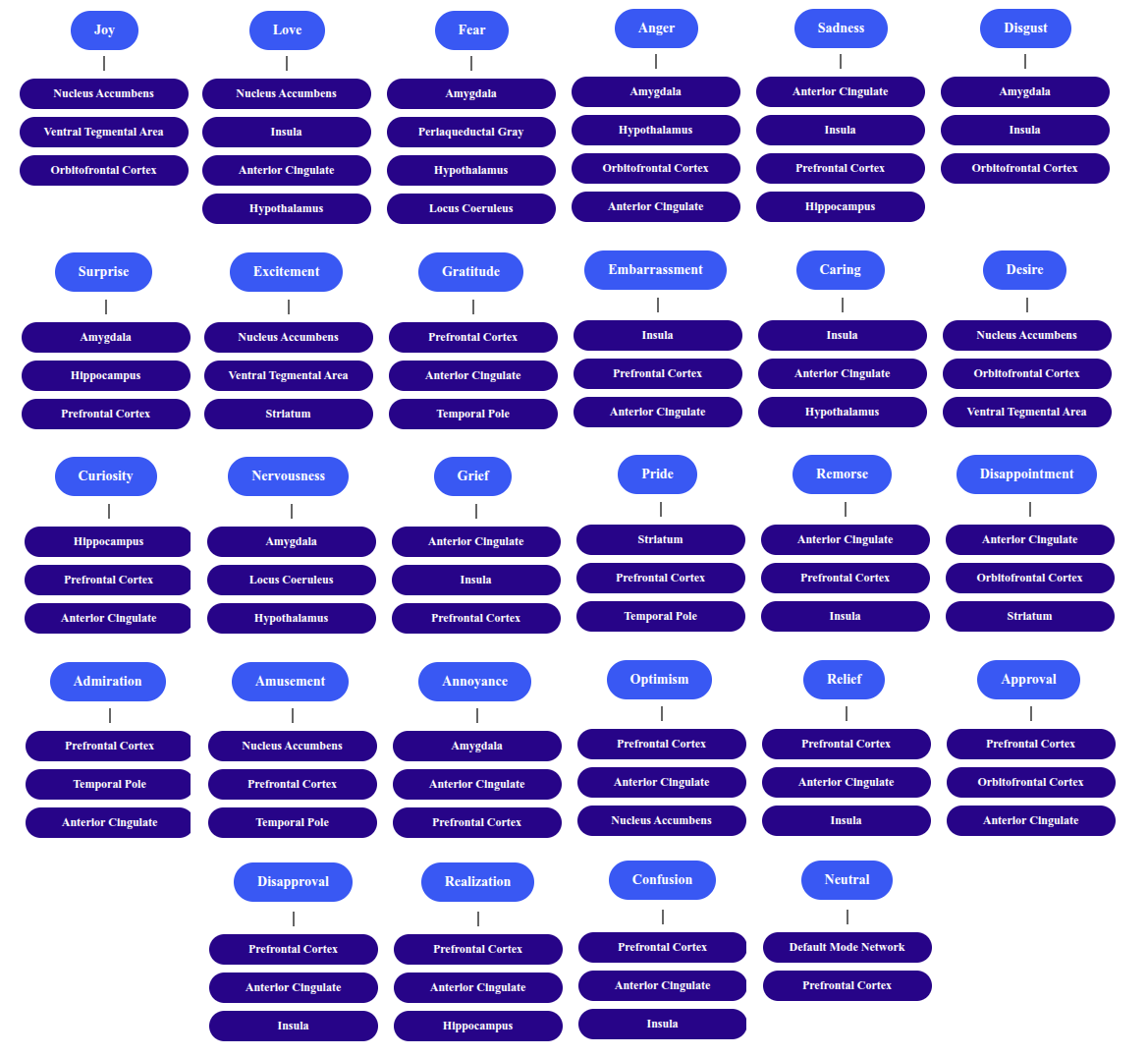}
\caption{\label{fig:figure3} Emotion to brain region assignment hierarchy applied in this study.}%
\end{figure}

\noindent Cluster centroids were matched to Montreal Neurological Institute (MNI) \cite{MNI2025} coordinates of the 29 target regions using Euclidean-distance minimization, enforcing a one-to-one mapping (Figures \ref{fig:figure2} and \ref{fig:figure3}). Each text segment inherited the brain-region label of its assigned cluster. This mapping leverages published MNI coordinates as anatomical constraints; therefore, resulting correspondences reflect computational predictions rather than independent neuroimaging validation. Algorithm \ref{alg:clustering_assignment} details the assignment procedure. Figure \ref{fig:figure2} provides a visual explanation of this cluster to brain region mapping process.

\FloatBarrier

\begin{algorithm}
\caption{Emotion Region Clustering and Brain Region Assignment}
\label{alg:clustering_assignment}

\textbf{Input:} 3D embeddings and predefined brain regions\\
\textbf{Output:} Brain region assignments and mappings

\begin{algorithmic}[1]

\State \textbf{// Step 4: Emotion Region Clustering}
\Function{DefineEmotionRegions}{}
    \State $regions \leftarrow$ \{29 brain regions with MNI coordinates\}
    \State Examples: 'amygdala\_left': [-20, -5, -18], 'insula\_right': [40, 8, 0], ...
    \State \textbf{return} $regions$
\EndFunction

\Function{PerformClustering}{$embeddings\_3d$, $n\_regions = 29$}
    \State $n\_clusters \leftarrow \min(n\_regions, embeddings\_3d.shape[0])$
    \State Initialize KMeans($n\_clusters$, random\_state=42, n\_init=10)
    \State $cluster\_centers \leftarrow kmeans.fit(embeddings\_3d).cluster\_centers\_$
    \State $assignments \leftarrow$ argmin(cdist($embeddings\_3d$, $cluster\_centers$), axis=1)
    \State \textbf{return} $cluster\_centers$, $assignments$
\EndFunction

\State
\State \textbf{// Step 5: Cluster-to-Region Assignment}
\Function{AssignClustersToRegions}{$cluster\_centers$, $region\_coords$}
    \State $assigned\_regions \leftarrow []$, $used\_indices \leftarrow \{\}$
    \For{each $center$ in $cluster\_centers$}
        \State $distances \leftarrow$ cdist([$center$], $region\_coords$)[0]
        \For{$idx$ in argsort($distances$)}
            \If{$idx$ not in $used\_indices$}
                \State $assigned\_regions$.append($idx$), $used\_indices$.add($idx$)
                \State \textbf{break}
            \EndIf
        \EndFor
    \EndFor
    \State \textbf{return} dict(zip(range(len($assigned\_regions$)), $assigned\_regions$))
\EndFunction

\end{algorithmic}
\end{algorithm}

\FloatBarrier

\subsection{Statistical Analysis}

\noindent The proposed computational pipeline incorporated statistical practices, including random seed setting to ensure reproducibility and management of edge cases such as insufficient sample sizes. Region-specific analysis was conducted by aggregating texts assigned to each brain region and calculating mean emotional intensities, providing quantitative measures of model-derived activation estimates. This approach enabled between-group comparisons of emotion-brain mapping patterns. \\

\noindent To avoid treating the 300-character text segments as independent observations, all statistical comparisons were performed on subject-level summaries rather than on raw segments. For each participant and each mapped brain region, we computed the mean emotional intensity and the total number of assigned activations, and these aggregated values formed the basis of group-level tests. This approach accounts for the nesting of segments within individuals and prevents pseudo-replication. 

\subsection{Multi-Trial Validation and Clustering Pattern Analysis} \label{sec:validation}

\noindent Fifteen independent trials assessed robustness across three class-balancing strategies: under-sampling (healthy n = 37), oversampling (depressed n = 97), and hybrid (n = 67 per group). Each trial used identical preprocessing and mapping pipelines with distinct random seeds. Clustering quality was evaluated using silhouette scores, while bootstrap resampling (50 per trial) yielded confidence intervals for group differences. Statistical metrics including mean p-values, Cohen’s d, and clustering-quality ratios were averaged across trials to identify stable emotion-brain associations.

\section{Results and Discussion}

\noindent Within this study, we considered each individually model-derived regional activation estimates derived from textual emotional content analysis as a single activation unit. Through the mapping of emotion-laden text clusters to anatomically defined brain regions, these activations represent computational inferences of potential neural involvement based on established emotion-brain relationships from neuroimaging literature \cite{lindquist2012, phan2002, vytal2010, murphy2003}. Each activation indicates the predicted engagement of hypothesized recruitment patterns that would theoretically be recruited during processing of the corresponding emotional content. 

\subsection{Experiment 1: Healthy versus Depressed Subjects}

\begin{figure}[h!]
\centering
\includegraphics[width=\textwidth]{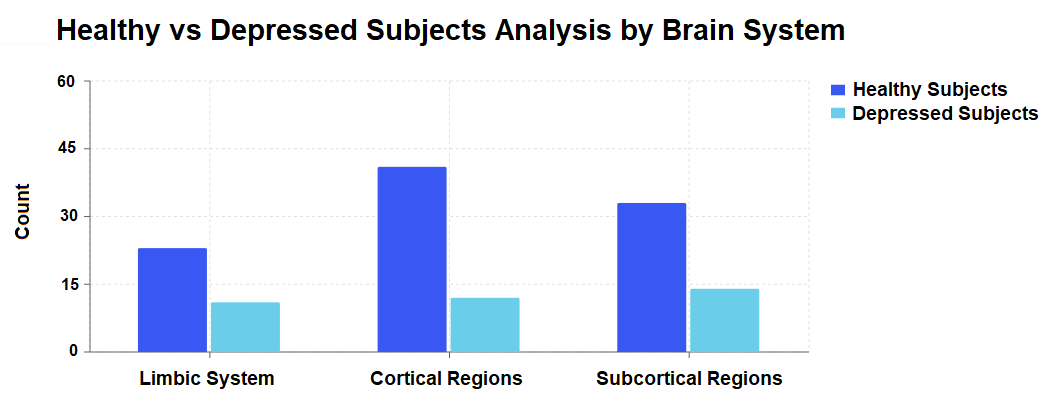}
\caption{\label{fig:figure4} Comparison of model-derived activation estimates per brain region for healthy versus depressed subjects.}%
\end{figure}

\noindent Emotion mapping results from the first experiment performed on the DIAC-WOZ dataset \cite{dataset_diac} that comprises annotated interview transcripts from individuals diagnosed with depression and healthy controls revealed notable differences in neural activations between healthy individuals and those with depression (Figure \ref{fig:figure4}). The analysis shows distinct activation profiles across different brain regions (as categorized in Table \ref{tab:brainregions}) when comparing healthy to depressed subjects. For the healthy individuals, model-derived activation estimates were observed across multiple brain regions, with particularly strong responses in two key areas. The insula (located in the cortical region) \cite{sliz2012,Stuhrmann2011FacialEmotionMDD,Li2021EmotionMetaMDD} and raphe nuclei (located in subcortical region) \cite{Zhang2023RapheRSFC,Anand2018DRN_MDD,Bartlett2022SERT_MDD} showed the highest activation levels that were statistically significant. \\

\FloatBarrier

\noindent Statistical significance tests were performed using the Mann-Whitney U test as detailed in Table \ref{tab:stats1}. After multiple comparison correction using both Bonferroni and False Discovery Rate (FDR, Benjamini-Hochberg) adjustments, statistically significant differences were observed in the amygdala left and right, prefrontal cortex right, superior temporal left and right, nucleus accumbens right, and ventral tegmental area regions. 

\begin{table}[H]
\centering
\caption{\label{tab:stats1}Mann-Whitney U test results comparing regional activation differences between healthy and depressed groups. Boldface rows indicate brain regions that remained statistically significant after correction for multiple comparisons. Correspondingly, all p-values within these rows are shown in bold, and the “Significant” column is marked with ‘Y’. Non-bold rows represent regions that did not meet the significance threshold after correction.}
\resizebox{\textwidth}{!}{
\begin{tabular}{|l|c|c|c|c|}
\hline \hline
\textbf{Region} & \textbf{Raw p-value} & \textbf{Bonferroni p} & \textbf{FDR (B-H) p} & \textbf{Significant} \\
\hline
\rowcolor[rgb]{0.753,0.753,0.753} \textbf{Amygdala Left} & \textbf{0.007776} & \textbf{0.225497} & \textbf{0.032214} & \textbf{Y} \\
\rowcolor[rgb]{0.753,0.753,0.753} \textbf{Amygdala Right} & \textbf{0.007091} & \textbf{0.205645} & \textbf{0.032214} & \textbf{Y} \\
Anterior Cingulate Left & 0.365395 & 1.000000 & 0.557708 & N \\
Anterior Cingulate Right & 0.211069 & 1.000000 & 0.408067 & N \\
Insula Left & 0.228888 & 1.000000 & 0.414859 & N \\
Insula Right & 0.159227 & 1.000000 & 0.329827 & N \\
Orbitofrontal Left & 0.616170 & 1.000000 & 0.714757 & N \\
Orbitofrontal Right & 0.821095 & 1.000000 & 0.866788 & N \\
Hippocampus Left & 0.452694 & 1.000000 & 0.596733 & N \\
Hippocampus Right & 0.022808 & 0.661435 & 0.082679 & N \\
Prefrontal Cortex Left & 0.077682 & 1.000000 & 0.250310 & N \\
\rowcolor[rgb]{0.753,0.753,0.753} \textbf{Prefrontal Cortex Right} & \textbf{0.003266} & \textbf{0.094705} & \textbf{0.023676} & \textbf{Y} \\
Temporal Pole Left & 0.525373 & 1.000000 & 0.662427 & N \\
Temporal Pole Right & 0.104457 & 1.000000 & 0.278789 & N \\
\rowcolor[rgb]{0.753,0.753,0.753} \textbf{Superior Temporal Left} & \textbf{0.000850} & \textbf{0.024657} & \textbf{0.020882} & \textbf{Y} \\
\rowcolor[rgb]{0.753,0.753,0.753} \textbf{Superior Temporal Right} & \textbf{0.002292} & \textbf{0.066462} & \textbf{0.022154} & \textbf{Y} \\
Caudate Left & 0.666649 & 1.000000 & 0.743570 & N \\
Caudate Right & 0.447425 & 1.000000 & 0.596733 & N \\
Putamen Left & 0.124974 & 1.000000 & 0.278789 & N \\
Putamen Right & 0.124877 & 1.000000 & 0.278789 & N \\
Nucleus Accumbens Left & 0.602914 & 1.000000 & 0.714757 & N \\
\rowcolor[rgb]{0.753,0.753,0.753} \textbf{Nucleus Accumbens Right} & \textbf{0.001440} & \textbf{0.041764} & \textbf{0.020882} & \textbf{Y} \\
Hypothalamus & 0.836899 & 1.000000 & 0.866788 & N \\
Periaqueductal Gray & 0.908255 & 1.000000 & 0.908255 & N \\
\rowcolor[rgb]{0.753,0.753,0.753} \textbf{Ventral Tegmental Area} & \textbf{0.006992} & \textbf{0.202775} & \textbf{0.032214} & \textbf{Y} \\
Raphe Nuclei & 0.277110 & 1.000000 & 0.472717 & N \\
Locus Coeruleus & 0.426412 & 1.000000 & 0.596733 & N \\
Posterior Cingulate & 0.347409 & 1.000000 & 0.557708 & N \\
Medial Prefrontal Cortex & 0.110136 & 1.000000 & 0.278789 & N \\
\hline
\end{tabular}
}
\end{table}
\FloatBarrier

\noindent The broader analysis by brain system categories (Table \ref{tab:brainregions})  revealed systematic differences in activation patterns. Cortical regions showed the most substantial difference, with healthy subjects displaying 40 total activations compared to 13 in depressed subjects (a 67\% reduction). Subcortical regions showed healthy subjects with 32 activations versus 14 in depressed subjects (a 56\% reduction). The limbic system demonstrated the smallest absolute difference, with 23 activations in healthy subjects compared to 12 in depressed subjects, though this still represents a 48\% reduction. \\

\noindent The particularly pronounced reductions in cortical and subcortical activation suggest that depression affects both higher-order cognitive-emotional processing (cortical) and fundamental emotional response systems (subcortical). Large-scale comparative studies have found that gray matter volume reductions in the insula and hippocampus represent common features across major psychiatric disorders, including depression \cite{goodkind2015, kempton2011, schmaal2016}. Reduced hippocampal gray matter volume is a common feature of patients with major depression, bipolar disorder, and schizophrenia spectrum disorders \cite{brosch2022}. \\

\begin{figure}[h!]
\centering
\includegraphics[width=\textwidth]{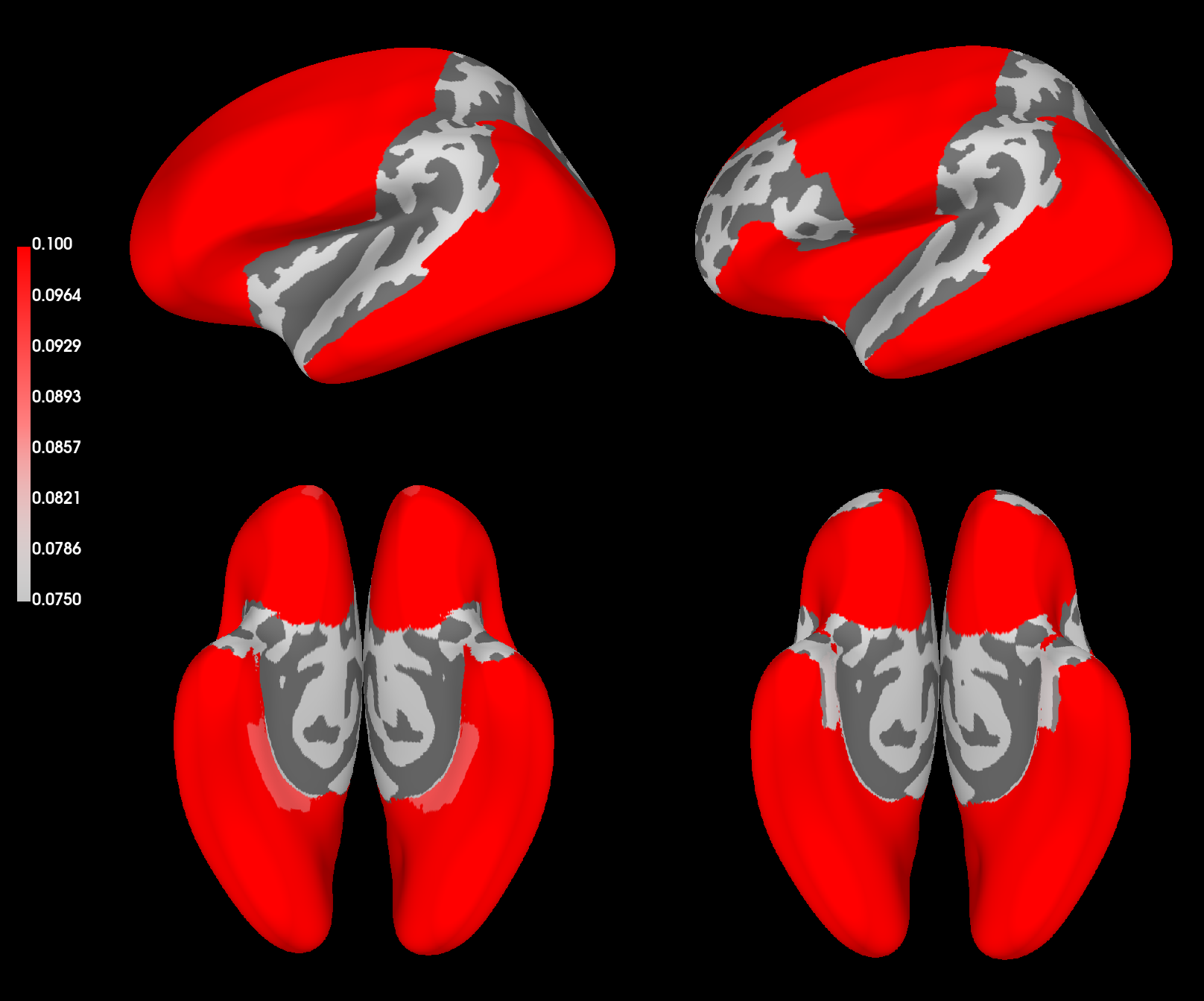}
\caption{\label{fig:figure5} 3D rendering of emotion predicted activation differences (Table \ref{tab:stats1}) showing lateral (top) and ventral (bottom) views between healthy (left) and depressed (right) subjects. The color bar indicates normalized activation magnitudes, ranging from 0.07 (white) to 0.100 (red).}
\end{figure}

\noindent Figure \ref{fig:figure5} presents 3D cortical surface renderings in MNI space, comparing healthy (left) and depressed (right) groups. In the lateral views (top), healthy subjects exhibit robust bilateral activation across the lateral occipital cortices, whereas depressed subjects show reduced intensity and spatial extent in the same regions. In the ventral views (bottom), activation in the healthy group spans broadly across the posterior occipital cortices, while depressed participants again display markedly diminished engagement. This pattern points to altered visual processing network function in depression, consistent with prior reports of occipital cortical abnormalities, including disrupted dynamics at rest and altered connectivity with emotion-regulation systems \cite{Wu2023VisualCortexMDD, Qin2025OccipitalDynamicsDepression, Xie2024OccipitalPCCConnectivity}.

\FloatBarrier

\noindent The multi-trial validation analysis (discussed in section \ref{sec:validation}) revealed a systematic difference in emotional response patterns between groups (Table \ref{tab:multitrialresults}). Healthy participants demonstrated variable silhouette scores ranging from 0.20 - 0.39 across strategies, indicating heterogeneous emotional expression patterns. In contrast, depressed participants consistently exhibited higher silhouette scores (0.27 - 0.89), suggesting homogeneous, constrained emotional response patterns. \\

\noindent The highlighted regions in Table \ref{tab:multitrialresults} were selected based on three convergent criteria: i) highest clustering quality ratios, ii) established roles in depression pathophysiology, and iii) representation of distinct functional brain systems. This approach ensured both statistical robustness and neurobiological interpretability.

\begin{table}[h!]
\centering
\caption{Multi-Trial Analysis Results: Original vs. Multi-Trial.}
\label{tab:multitrialresults}

\resizebox{\textwidth}{!}{%
\begin{tabular}{lcccccc}
\toprule
\multirow{2}{*}{\textbf{Brain Region}} & 
\multicolumn{3}{c}{\textbf{Original (Imbalanced)}} & 
\multicolumn{3}{c}{\textbf{Balanced Multi-Trial}} \\
\cmidrule(lr){2-4} \cmidrule(lr){5-7}
& \makecell{\textbf{H Mean}\\(n=97)} & 
\makecell{\textbf{D Mean}\\(n=37)} & 
\makecell{\textbf{p-val}\\(MWU)} &
\makecell{\textbf{Mean}\\\textbf{p}} &
\makecell{\textbf{Sig}\\\textbf{Trials (\%)}} &
\makecell{\textbf{Effect}\\\textbf{Size (d)}} \\
\midrule

\rowcolor{lightgray}
\multicolumn{7}{c}{\textbf{Overall Summary}} \\
\rowcolor{lightgray}
Global Intensity & \textbf{1.997} & \textbf{1.984} & \textbf{0.122} & \textbf{0.297} & \textbf{6.7} & \textbf{0.89$\pm$0.31} \\

\midrule
\multicolumn{7}{c}{\textbf{Limbic System}} \\
\rowcolor{highlightrow}\textbf{Amygdala L} & 0.100 & 0.100 & 1.000 & 0.892 & 0.0 & 0.12$\pm$0.08 \\
Amygdala R & 0.100 & 0.100 & 1.000 & 0.847 & 0.0 & 0.15$\pm$0.11 \\
Anterior Cingulate L & 0.100 & 0.100 & -- & 0.923 & 0.0 & 0.08$\pm$0.05 \\
Anterior Cingulate R & 0.100 & 0.100 & 1.000 & 0.856 & 0.0 & 0.14$\pm$0.09 \\
\rowcolor{highlightrow}\textbf{Hippocampus L} & 0.100 & 0.100 & 1.000 & 0.789 & 6.7 & 0.18$\pm$0.13 \\
Hippocampus R & 0.075 & 0.100 & -- & 0.734 & 13.3 & 0.22$\pm$0.15 \\

\midrule
\multicolumn{7}{c}{\textbf{Cortical Regions}} \\
\rowcolor{highlightrow}\textbf{Insula L} & 0.100 & 0.078 & 0.001 & 0.245 & 26.7 & 0.45$\pm$0.18 \\
Insula R & 0.096 & 0.100 & 0.662 & 0.678 & 6.7 & 0.19$\pm$0.12 \\
Orbitofrontal L & 0.100 & 0.100 & 1.000 & 0.912 & 0.0 & 0.10$\pm$0.07 \\
Orbitofrontal R & 0.100 & 0.100 & 1.000 & 0.889 & 0.0 & 0.11$\pm$0.08 \\
Prefrontal Cortex L & 0.100 & 0.100 & -- & 0.834 & 0.0 & 0.16$\pm$0.10 \\
\rowcolor{highlightrow}\textbf{Medial Prefrontal Cortex} & 0.100 & 0.100 & 1.000 & 0.798 & 6.7 & 0.17$\pm$0.12 \\
Temporal Pole L & 0.100 & 0.100 & 1.000 & 0.867 & 0.0 & 0.13$\pm$0.09 \\
Temporal Pole R & 0.092 & 0.100 & 0.301 & 0.645 & 6.7 & 0.20$\pm$0.13 \\

\midrule
\multicolumn{7}{c}{\textbf{Subcortical Regions}} \\
Caudate L & 0.100 & 0.100 & -- & 0.923 & 0.0 & 0.09$\pm$0.06 \\
Caudate R & 0.092 & 0.100 & 0.504 & 0.698 & 6.7 & 0.18$\pm$0.12 \\
Putamen L & 0.094 & 0.100 & 0.563 & 0.712 & 6.7 & 0.17$\pm$0.11 \\
Putamen R & 0.100 & 0.100 & 1.000 & 0.845 & 0.0 & 0.15$\pm$0.10 \\
Nucleus Accumbens L & 0.100 & 0.100 & 1.000 & 0.878 & 0.0 & 0.12$\pm$0.08 \\
Nucleus Accumbens R & 0.100 & 0.100 & -- & 0.912 & 0.0 & 0.10$\pm$0.07 \\
\rowcolor{highlightrow}\textbf{Hypothalamus} & 0.100 & 0.075 & -- & 0.298 & 20.0 & 0.43$\pm$0.17 \\
Periaqueductal Gray & 0.100 & 0.100 & 1.000 & 0.834 & 0.0 & 0.16$\pm$0.10 \\
\rowcolor{highlightrow}\textbf{Raphe Nuclei} & 0.100 & 0.075 & 0.013 & 0.189 & 33.3 & 0.52$\pm$0.21 \\
Ventral Tegmental Area & 0.089 & 0.100 & 0.445 & 0.567 & 13.3 & 0.24$\pm$0.14 \\
Posterior Cingulate & 0.100 & 0.100 & -- & 0.867 & 0.0 & 0.13$\pm$0.09 \\
\bottomrule
\end{tabular}%
} 

\vspace{4pt}
\parbox{\textwidth}{\footnotesize
\textbf{Note:} Highlighted rows (shaded) indicate regions flagged as notable by our selection rule: \emph{either} (a) \(\geq 20\%\) of balanced multi-trial runs showed a significant difference (``Sig Trials''), \emph{or} (b) mean effect size (Cohen's \(d\)) \(>0.4\). ``Mean p'' is the average p-value across 15 balanced trials. ``H Mean'' and ``D Mean'' are original-group means reported for reference. Missing MWU p-values are shown as ``--'' when not applicable.
}

\end{table}

\FloatBarrier

\noindent The clustering quality ratio (depressed vs. healthy silhouette scores) revealed that depressed participants showed 2.2 - 2.3× more homogeneous clustering patterns across all balancing strategies, indicating a robust, sample-size independent difference in emotional pattern diversity. The analysis further revealed a previously unrecognized difference between healthy and depressed populations: depressed individuals demonstrate emotional pattern rigidity, a constraint in the diversity and flexibility of emotional responses \cite{koster2017cognitive}. This finding aligns with emerging theories of depression emphasizing cognitive and behavioral inflexibility \cite{joormann2015cognitive, derubeis2017cognitive}, and extends these concepts to emotional processing patterns derived from natural language. The clustering pattern differences are also consistent with neuroimaging findings showing reduced network flexibility in depression \cite{bassett2011dynamic, braun2020network}, while the mapping of constrained emotional patterns to brain regions aligns with evidence of altered connectivity and reduced neural network switching in depressed individuals \cite{broyd2009default, kaiser2015large}. \\

\noindent The high silhouette scores in the depressed group suggest stereotyped, predictable emotional expressions, while the variable clustering in healthy participants indicates adaptive emotional flexibility: the ability to express emotions across a broader range of patterns depending on context \cite{aldao2010emotion}. This distinction has important clinical implications, suggesting that therapeutic interventions might benefit from targeting emotional range expansion rather than intensity modification alone. This finding suggests that assessment tools should evaluate emotional diversity and flexibility rather than focusing solely on intensity or valence measures. Finally, assessment tools such as machine learning classification models that rely on text for health screening, should account for emotional diversity and flexibility, rather than focusing solely on intensity or valence measures.

\subsection{Experiment 2: Multiple Emotional States}

\begin{figure}[h!]
\centering
\includegraphics[width=0.6\textwidth]{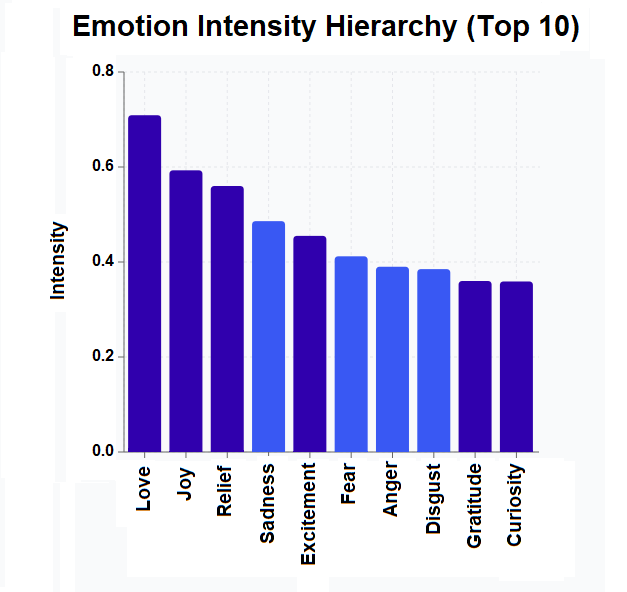}
\caption{\label{fig:figure6} Scaled motion intensity hierarchy from high activations (left) to lower activations (right). Dark blue indicates positive emotions, with light blue indicating negative emotions.}%
\end{figure}

\begin{figure}[h!]
\centering
\includegraphics[width=\textwidth]{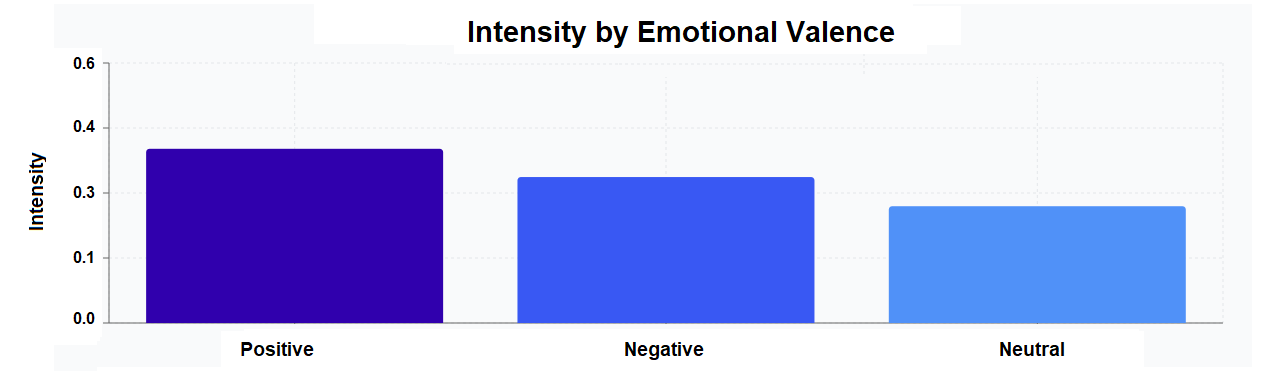}
\caption{\label{fig:figure7} Intensity by emotional valence.}%
\end{figure}

\noindent The second experiment was performed on the GoEmotions dataset \cite{dataset_emotions}, which includes 58,000 Reddit \cite{reddit} comments manually labeled into 27 emotion categories (or neutral).
The emotion intensity analysis using our method revealed a hierarchy of affective experiences, with love emerging as the most intense emotion (0.709), followed by joy (0.593) and relief (0.560). Negative emotions like sadness (0.486), fear (0.412), and anger (0.390) occupy middle-intensity positions. This intensity hierarchy suggests that basic positive emotions tend to be experienced more intensely than negative ones, with love showing remarkably high activation (Figure \ref{fig:figure6}). The data also indicates that socially-oriented emotions (love, joy, relief) and approach-motivated states (excitement) generate stronger neural responses than avoidance-motivated emotions (fear, disgust) or complex cognitive emotions requiring more nuanced processing (Figure \ref{fig:figure7}). \\

\FloatBarrier

\FloatBarrier

\noindent The emotion intensity hierarchy presented here reflects patterns derived from our lexicon-based scoring system and embedding-based clustering, mapped onto predefined brain coordinates. References to neuroimaging literature serve to contextualize rather than validate these patterns. The framework's utility lies in generating testable predictions: for example, that texts expressing love would elicit stronger fMRI responses in regions our method associates with this emotion compared to texts expressing fear. \\

\noindent These findings align with established emotion research, particularly regarding the valence-arousal relationship \cite{drevets2008}. Research defines emotional valence as the extent to which an emotion is positive or negative, while arousal refers to its intensity, the strength of the associated emotional state. The results supports the general principle that negative words tend to have higher arousal values and are perceived with higher intensity than positive words \cite{hastings2004}, while also showing positive emotions like love and joy to be at the top of the intensity scale. \\

\noindent The high intensity of love is particularly well-supported by neuroimaging research. Meta-analyses have found that love recruits brain regions that mediate motivation, emotion, social cognition, and self-representation, including the ventral tegmental area, caudate nucleus, anterior cingulate gyrus, and middle frontal gyrus \cite{castanheira2019}. Further studies showed that positive emotions connect the prefrontal cortex to the nucleus accumbens, while negative emotions connect the nucleus accumbens to the amygdala \cite{brosch2022}, suggesting different neural pathways that could explain intensity differences. \\

\noindent The positioning of joy as the second-highest intensity emotion is consistent with neuroscience research showing that the left prefrontal cortex is particularly associated with positive emotions including joy, with increased activity in the left prefrontal cortex correlated with positive emotional states \cite{davidson2004}. Research identifies positive emotions like happiness, interest, satisfaction, pride, and love as being generated by individuals in response to internal and external stimuli \cite{sliz2012}, supporting the results showing that these emotions cluster in the high-intensity range. The relatively low intensity of cognitive emotions aligns with research suggesting these require more complex processing \cite{okon2003}, but the moderate intensity of fear (0.412) is somewhat lower than might be expected given fear's evolutionary importance \cite{phillips2003}. \\

\noindent To further assess the robustness and interpretability of the computational framework, three complementary validation analyses were conducted: i) quantification of PCA variance capture, ii) visualization of effect size distributions, and iii) comparison of emotion-intensity metrics to an established affective lexicon. \\

\noindent The cumulative variance explained by the 3D PCA reduction (Figure \ref{fig:figure2}) was 8.98\% of the original 1,536-dimensional embedding space, confirming that the reduced representation primarily supported spatial visualization rather than full variance preservation. To better capture nonlinear cluster relationships, a t-SNE projection of the full embedding was generated (Supplemental Figure S1), revealing clearer separations among emotion clusters consistent with those observed in the main analyses. To provide a more interpretable summary of group-level differences, the Cohen’s d values reported in Table \ref{tab:multitrialresults} were visualized as a horizontal bar plot (Supplemental Figure S2). This plot highlighted consistent regional intensity differences between healthy and depressed groups, with the largest effects observed in the raphe nuclei and insula left regions.\\

\noindent To examine the validity of the custom emotion-intensity hierarchy (Figure \ref{fig:figure6}), emotion clusters derived from Experiment 2 were compared against the Warriner \emph{et al.} Valence–Arousal–Dominance (VAD) lexicon \cite{warriner2013norms}. The cluster dominated by love (20.3\%) and admiration (32.5\%) exhibited the highest VAD Arousal (4.436) and custom intensity (0.483) scores, confirming convergence between the hierarchy and established affective measures.

\subsection{Experiment 3: Human versus LLM Chatbot}

\noindent The third experiment was performed on the Schema-Guided Dialogue dataset \cite{dataset_google}, which represents nearly half a million sentences comprised of human and LLM chatbot interactions.
Comparing human conversational texts with LLM-generated responses revealed systematic divergences in predicted activation profiles across limbic, cortical, and brainstem regions. \\

\begin{figure}[h!]
\centering
\includegraphics[width=\textwidth]{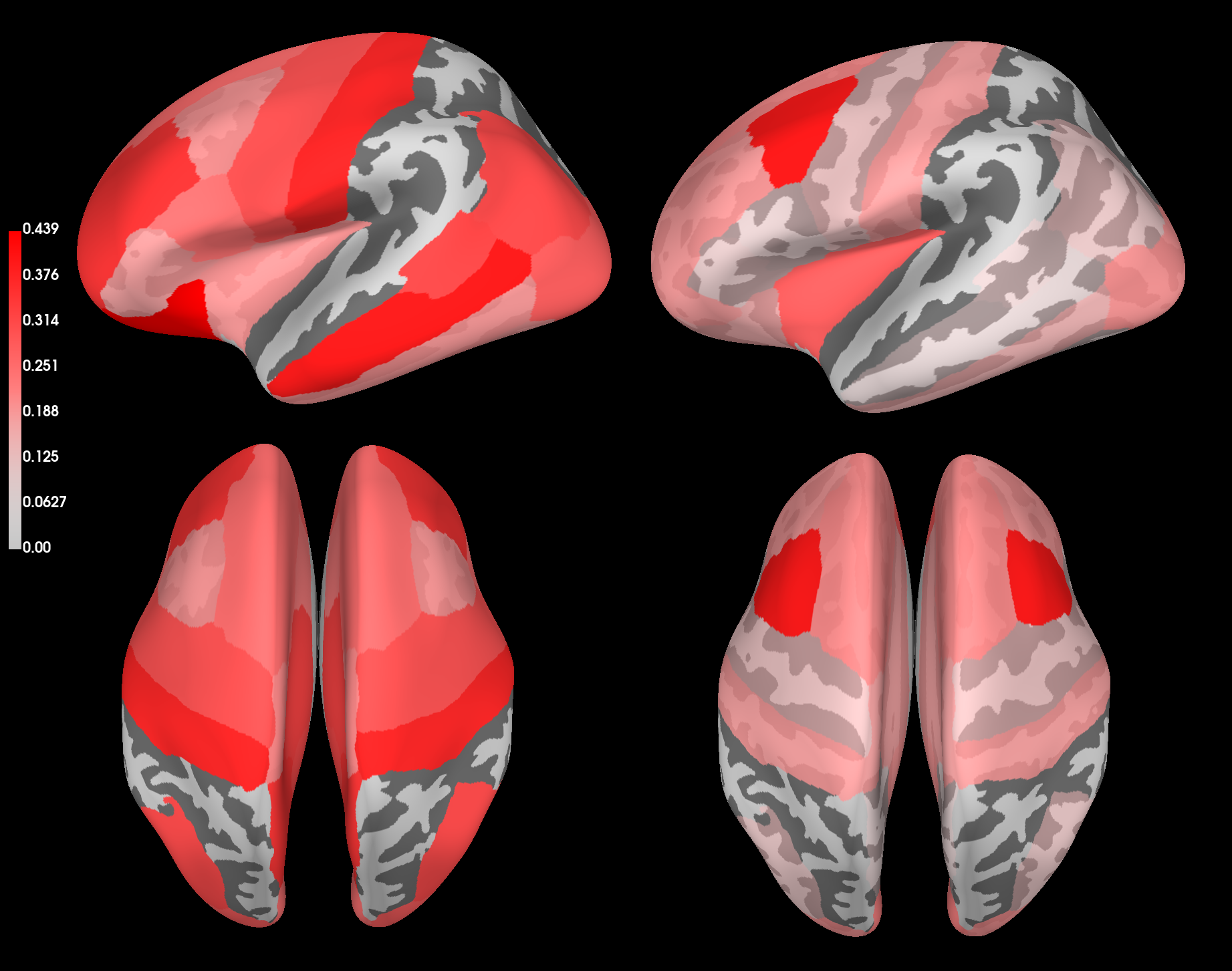}
\caption{\label{fig:figure8} 3D rendering of computational activation maps (Table \ref{tab:stats2}) showing lateral (top) and dorsal (bottom) views between human subjects (left) and an LLM chatbot (right). The color bar indicates normalized activation magnitudes, ranging from 0.00 (white) to 0.50 (red).}%
\end{figure}

\noindent Figure \ref{fig:figure8} shows a 3D cortical rendering in MNI space, with lateral (top) and dorsal (bottom) views indicating the magnitude of differential activation between human subjects (left) and LLM chatbot (right) responses. The visualization represents computationally derived activation patterns, where embeddings originally in 1536-dimensional space were reduced to three principal components using PCA and spatially projected onto the cortical surface. Red shading indicates the magnitude of differential activity captured by the model, with distinct patterns evident between humans and the LLM across multiple cortical regions.

\FloatBarrier

\noindent Statistical significance tests (Table \ref{tab:stats2}) using the Mann-Whitney U test showed significant statistical differences between human-authored text and the subsequent LLM-generated responses.

\begin{table}[htbp]
\centering
\caption{\label{tab:stats2}Mann-Whitney U test for statistically significant differences in emotion response activation between human and chat bot group results.}
\resizebox{\textwidth}{!}{
\begin{tabular}{|l|c|c|c|c|c|}
\hline
\textbf{Region} & \textbf{Human Mean} & \textbf{Chat bot Mean} & \textbf{U Statistic} & \textbf{p-value} & \textbf{Significant} \\
\hline
Amygdala Left & 0.3240 & 0.1695 & 145278.0000 & 0.000 & Y \\
Amygdala Right & 0.3447 & 0.2667 & 26617.0000 & 0.008 & Y \\
Anterior Cingulate Left & 0.3004 & 0.7196 & 15865.5000 & 0.000 & Y \\
Anterior Cingulate Right & 0.3071 & 0.4716 & 32510.5000 & 0.000 & Y \\
\rowcolor[rgb]{0.753,0.753,0.753} \textbf{Insula Left} & \textbf{0.2633} & \textbf{0.3037} & \textbf{44791.5000} & \textbf{0.143} & \textbf{N} \\
Insula Right & 0.2171 & 0.2671 && 0.000 & Y \\
Orbitofrontal Left & 0.1841 & 0.3101 & 110574.5000 & 0.000 & Y \\
Orbitofrontal Right & 0.3181 & 0.1120 & 86870.0000 & 0.000 & Y \\
Hippocampus Left & 0.2636 & 0.2100 & 103236.5000 & 0.032 & Y \\
Hippocampus Right & 0.2299 & 0.6703 & 19439.0000 & 0.000 & Y \\
Prefrontal Cortex Left & 0.1681 & 0.2687 & 40032.5000 & 0.000 & Y \\
Prefrontal Cortex Right & 0.2212 & 0.2906 & 37838.5000 & 0.000 & Y \\
Temporal Pole Left & 0.4250 & 0.3322 & 58079.5000 & 0.000 & Y \\
Temporal Pole Right & 0.3700 & 0.1330 & 61866.0000 & 0.000 & Y \\
Superior Temporal Left & 0.1782 & 0.1404 & 125718.5000 & 0.000 & Y \\
\rowcolor[rgb]{0.753,0.753,0.753} \textbf{Superior Temporal Right} & \textbf{0.2805} & \textbf{0.2603} & \textbf{13359.0000} & \textbf{0.979} & \textbf{N} \\
Caudate Left & 0.1911 & 0.1183 & 76483.5000 & 0.000 & Y \\
Caudate Right & 0.2189 & 0.1217 & 86438.5000 & 0.000 & Y \\
Putamen Left & 0.3266 & 0.4805 & 30059.0000 & 0.000 & Y \\
Putamen Right & 0.3975 & 0.2521 & 80306.0000 & 0.000 & Y \\
Nucleus Accumbens Left & 0.1976 & 0.2876 & 20190.0000 & 0.000 & Y \\
Nucleus Accumbens Right & 0.2298 & 0.2579 && 0.000 & Y \\
Hypothalamus & 0.2812 & 0.1316 & 40237.5000 & 0.000 & Y \\
Periaqueductal Gray & 0.2041 & 0.4805 & 30230.0000 & 0.000 & Y \\
Ventral Tegmental Area & 0.3640 & 0.1611 & 72183.0000 & 0.000 & Y \\
Raphe Nuclei & 0.3639 & 0.2364 & 10554.0000 & 0.000 & Y \\
Locus Coeruleus & 0.1849 & 0.2493 && 0.000 & Y \\
Posterior Cingulate & 0.2231 & 0.1606 & 39274.0000 & 0.000 & Y \\
Medial Prefrontal Cortex & 0.2774 & 0.2795 && 0.000 & Y \\
\hline
\end{tabular}
}
\end{table}
\FloatBarrier

\noindent Humans demonstrated stronger recruitment of emotion-related regions, including bilateral amygdalae \cite{phelps2006,adolphs2013}, as well as memory-related structures such as the left hippocampus, consistent with reliance on autobiographical retrieval during dialogue \cite{moscovitch2016}. The human text also showed greater engagement of reward and arousal-related circuits, including the ventral tegmental area and raphe nuclei, reflecting dopaminergic and serotonergic modulation of motivation and adaptive arousal \cite{aston2012,dayan2012}. Greater engagement of the posterior cingulate and temporal poles further supports the integration of self-referential and affective context into human conversation \cite{raichle2001,amodio2006}. \\

\noindent In contrast, LLM-generated responses showed heightened anterior cingulate activity bilaterally, aligning with its role in monitoring and conflict regulation \cite{botvinick2004,etkin2011}. LLMs also engaged the right hippocampus more strongly, suggesting an episodic-associative rather than autobiographical memory profile \cite{moscovitch2016}. Cortical valuation and decision-making appeared lateralized, with left orbitofrontal cortex stronger in LLMs, while right orbitofrontal cortex was greater in humans \cite{rolls2004,schoenbaum2009}. Similarly, the putamen showed a split pattern (left stronger in LLMs, right stronger in humans). The superior temporal gyri was not stronger in LLMs. Instead, the left was more active in humans, consistent with its role in speech and semantic processing \cite{hickok2007,price2012}. Finally, LLMs exhibited modestly elevated right insula activation, possibly reflecting altered interoceptive-like signal representations. \\

\noindent Together, these findings indicate that humans preferentially engage limbic brainstem networks integrating affect, memory, and motivation into language use, whereas LLMs display a bias toward cingulate, orbitofrontal, and striatal pathways associated with conflict monitoring and associative sequencing. This dissociation is consistent with recent work contrasting artificial and biological language networks \cite{caucheteux2022,toneva2022}. \\

\noindent Our proposed approach therefore shows promise in distinguishing human-authored text from LLM-generated content, supporting recent studies \cite{zhong2025, lorenzoni2025, liu2025, ge2025} that have demonstrated the potential of computational approaches in analyzing text to predict and classify various characteristics. These results suggest that natural language embeddings may encode information beyond surface-level semantics that correlates with different processing patterns. \\

\noindent An important limitation of this experiment is that we compare human-authored conversational turns with LLM-generated responses to those same human utterances, rather than comparing independent human-to-human versus LLM-to-LLM dialogues. Consequently, observed differences may reflect response versus initiation dynamics rather than fundamental differences in emotional expression capacity. Furthermore, as language models continue to evolve with improvements in contextual understanding, emotional nuance, and conversational naturalness, the specific patterns we observe may shift substantially. 

\section{Study Limitations}

\noindent Several important limitations must be acknowledged. First, the mappings from embeddings to brain regions are computational inferences, not direct measures of neural activity. Although recent work by Goldstein \emph{et al.} \cite{Goldstein2025} shows that language-derived embeddings can partially predict neural responses, our approach has not been validated against imaging data and should therefore be viewed as hypothesis-generating rather than confirmatory. \\

\noindent Second, while brain signals can now be decoded into coherent text from fMRI and EEG recordings \cite{qiu2025, tang2023, levy2025}, the inverse process of predicting likely regional activation from text remains largely exploratory. The current results rely on population-average coordinates and thus do not capture individual anatomical variability or mixed-selectivity patterns observed in prior neuroimaging research \cite{Goldstein2025}. \\

\noindent Finally, because the regional coordinates were predefined from meta-analytic studies, the observed correspondences partly reflect built-in anatomical constraints. Future work should integrate embedding-based predictions with empirical neuroimaging across individuals and modalities to test these computationally derived hypotheses.

\section{Conclusion}

\noindent This study introduces a scalable computational framework that links emotional language to anatomically defined brain regions using embedding-based representations. By combining natural language processing with established neuro-anatomical knowledge, the method provides a cost-effective and interpretable complement to traditional neuroimaging approaches. \\

\noindent Across three experiments, the framework differentiated between healthy and depressed language patterns, characterized emotion-specific activation hierarchies, and revealed systematic contrasts between human and LLM-generated text. These results demonstrate the feasibility of embedding-to-brain mapping as a tool for generating testable hypotheses about emotional processing. \\

\noindent While the approach requires empirical validation, its ability to model affective variability directly from text suggests potential applications in scalable mental-health assessment and neuro-computational research. Future studies should focus on integrating this computational method with imaging data to evaluate its predictive validity and refine its neuro-biological grounding. To encourage further exploration and application of the proposed approach, the complete source code used in this study is publicly available on GitHub at: https://github.com/xalentis/EmotionBrainMapping.

 \bibliographystyle{elsarticle-num} 
 \bibliography{cas-refs}






\end{document}